# AUTOMATING THE AUDIT OF ELECTRONIC INVOICES WITH A SOFTWARE ROBOT

[1] Tian Jun Cheng, [2] Chia Jung Chen, [3] Yao Lin Ong, [4] Yi Fang Yang, [5] Guang Yih Sheu*

[1] Chi Mei Medical Center, Tainan City 71004, Taiwan (R.O.C.)
[2] Department of Information Systems, Chi Mei Medical Center, Tainan City 71004, Taiwan (R.O.C.)
[3] Executive Master of Business Administration, Chang-Jung Christian University, Tainan City 711301, Taiwan (R.O.C.)
[4] Department of Accounting and Information Systems, Chang-Jung Christian University, Tainan City 711301, Taiwan (R.O.C.)
[5] Department of Innovation Application and Management/Accounting and Information Systems, Chang-Jung Christian University, Tainan City 711301, Taiwan (R.O.C.)

Taiwan's Chi Mei Medical Center has completed four challenges mentioned in published robotic process automation (RPA) researches including automating a dynamic process, designing feasible human-robot collaboration, incorporating with other emerging technologies, and bringing positive business impacts. Its executives called a committee to implement the electronic invoicing. This implementation includes the creation of a software robot to download automatically cloud electronic invoice (E-invoice) data from Taiwan's E-invoice platform and detect the inconsistency between them and on-premise data. This bot operates when internal auditors are off their office. They satisfied this software robot since the remaining work is only verifying the resulting inconsistency. The Chi Mei Medical Center measured the time and costs before and after adopting software robots to audit E-invoice; consequently, it welcomed more bots automating other business processes. In conclusion, integrating a software robot with other emerging technologies mitigates the possible errors provided by this bot. A good human-robot collaboration relies on the consideration of human perspective in choosing RPA tasks. Free bot creators are sufficient to verify that automating a business process using a bot is a reasonable investment.

## 1. Introduction

Taiwan's Chi Mei Medical Center is one of medical centers at Taiwan. However, it still faced the survival challenge. Observing existing errors in issuing paper invoices, the Chi Mei Medical Center asks its internal auditors to replace paper invoices with E-invoices. Paper invoices cause frauds such as undeclared invoices. It requires a long team of internal auditors for resolving those problems.

In Taiwan, uploading each E-invoice to Taiwan's E-invoice platform (https://www.einvoice.nat.gov.tw/) is necessary. Although this upload may be immediate or periodic, we may employ the E-invoice platform to detect frauds. The clues of frauds lie in the inconsistency between cloud E-invoice and on-premise data.

However, browsing Taiwan's E-invoice platform to detect inconsistencies between cloud E-invoice and on-premise data is still inefficient. It is easy to expect that internal auditors of the Chi Mei Medical Center will refuse to browse excessive cloud E-invoices on Taiwan's E-invoice platform. This browse is uninteresting and repetitive. Besides, although the Chi Mei medical center has adopted a commercial any commercial Enterprise Resources Planning (ERP) system to manage accounting data, it doesn't have the function of connecting Taiwan's E-invoice platform. This platform is unique.

We may employ recent free bot creators such as UiPath (https://www.uipath.com/) (community version) to create software robots for automating the browsing of Taiwan's E-invoice platform. The IEEE Standard Association defines a software robot as [1] "A preconfigured software instance that uses business rules and predefined activity choreography to complete the autonomous execution of a combination of processes, activities, transactions, and tasks in one or more unrelated software systems to deliver a result of service with human exception management." A software robot may reproduce the work that humans do. It may be called a software robot or a bot.

Since software robots are one of the young information technologies. Even executives of the Chi Mei Medical Center were afraid that their internal auditors can't accept the RPA technology. However, recent bot creators are easy enough to let employees of the Chi Mei Medical Center create a software robot by themselves. The resulting software robot can log automatically in Taiwan's E-invoice platform, download cloud E-invoice data, and detect inconsistencies between cloud E-invoice and on-premise data. Considering human-robot collaboration, a bot can operate when auditors get off their work. If the Chi-Mei medical center is successful in implementing electronic invoicing, it experiences denote a good industry-specific application for the software robot technology.

Nevertheless, an industry-specific application of software robots to automate an audit of E-invoices was rare. The Chi Mei Medical Center had no choice but to create its software robot by their own employees. This study presents the Chi Medical Center's development of a software robot to automate the electronic invoice auditing. We focus on the dynamic process automation, human-robot collaboration, RPA and other emerging technologies, tangible and intangible business impacts in this development.

The remainder of this study contains five sections. Sec. 2 presents a literature review relevant to this study. Sec. 3 illustrates how the Chi Mei Medical Center implemented electronic invoicing. Sec. 4 shows the effects of implementing this electronic invoicing. Based on this section, Sec. 5 presents some discussions. Sec. 6 offers the conclusion of this study.

## 2. Literature review

The Chi Mei Medical Center has asked the Chang Jung Christian University to provide some advices in creating a software robot to automate the audit of E-invoices. We searched those published studies presenting the procedures of coding such a bot and possible challenges encountered in implementing the procedures. However, only general procedures were available. It is easy find similar steps in a system analysis and design book (e.g. [2]). They provide elementary supports for the Chi Mei Medical Center in creating a software robot. For example, Huang and Vasarhelyi [3] proposed a framework illustrating the RPA in auditing. This framework contains four stages. The first stage is selecting procedures to be automated by a software robot. The second stage is modifying those procedures with considering the application of a bot. The third stage is creating a software robot according to results of the second stage. The final stage is operating the resulting bot and evaluating its performance.

Perdana, et al. [4] suggested RPA solutions for four scenarios and possible challenges. The third scenario may be the most like the audit of E-invoices. Its goal is detecting understated liabilities. If a software robot is available for automating this detection, auditors can only review a spreadsheet recorded detection results. Nevertheless, Perdana, et al. [3] thought software robots may output incorrect detections. For example, a bot may use the OCR to digitize a document. However, this digitization may result in errors. Auditors should manually correct these errors.

Except for the audit demands, industry-specific applications of software robots are available for many fields. For example, Telefónica O2, the second-largest telecommunications provider in the United Kingdom, had adopted over 100 bots to handle 500000 transactions per month [5]. A software robot was created to validate data, access databases, create documents, and upload the repository in the process of updating London Premium Advice Notes. Australia and New Zealand Group had built software robots to dramatically reduce the need of human workers [6].

## 3. Electronic invoicing

The Chi Mei Medical Center now has 13 administrative departments, including the Human Resources Department, the Accounting Office, and the Resources Office. The Accounting Office is responsible for accounting and audit systems. The accounting system creates the issuance of notes for incoming and outgoing, processes accounting affairs, and compiles financial statements. The audit system reviews incoming and outgoing accounting of all branches and accounts and collects receivables. To eliminate papers, improve the correctness and speed of operations, executives of the Chi Mei Medical Center established a cross-department digital transition team. The first job of this team was integrating the Director's Office, the Business Administration Department, the Materials Office, the Information Office, the Accounting Department, the Cashier Division, and the Pharmacy Department to introduce electronic invoicing in placing an order. They identified that placing an order has four steps:
1. Manufacturers of this order issue payment invoices. These invoices may be hand-written or hard copies of E-invoices. Fig. 1 exemplifies a hard copy of an E-invoice.

2. The Account Office of the Chi Mei Medical Center stores order information in its enterprise information planning or accounting information system.
3. If manufacturers are authorized traders of Taiwan's E-invoice platform, they should upload E-invoice data to this website. This upload may be immediate or periodical.
4. Internal auditors of the Chi Mei Medical Center review the consistency between invoices from manufacturers and its accounting data. This review of invoices may be one-by-one or in batches. In reviewing a hard copy of an E-invoice, an internal auditor can scan its QR code to browse E-invoice on Taiwan's E-invoice platform. If an audited invoice is paper-based, internal auditors have no choice but to use their eyes to audit this paper invoice.

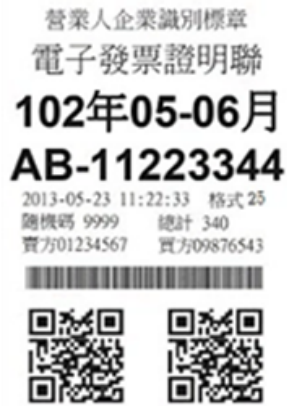

Fig. 1 A hard copy of an E-invoice (in traditional Chinese)

The Chi Mei Medical Center knew that the last step was the bottleneck in invoicing. Even if hard copies of E-invoices were adopted, browsing Taiwan's E-invoice platform to detect the inconsistency between cloud and on-premise data was still time-consuming. They may print cloud and on-premise data to facilitate the detection of inconsistency. However, it is easy to expect that internal auditors refuse to complete such detection since it impairs eyesight. Therefore, the Chi Mei Medical Center adopted UiPath to create a software robot to automate the step of detecting the inconsistency between cloud E-invoice and on-premise data. Fig. 2 illustrates the resulting software robot. This bot automatically logs in to Taiwan's E-invoice platform and imports cloud E-invoice data into an Excel file. It can wait for new E-invoices and import cloud E-invoice data into an Excel file. Detecting the inconsistency between cloud E-invoice and on-premise data is next implemented in this Excel file. A separate module was employed to output the inconsistency. Internal auditors only need to confirm these outputs.

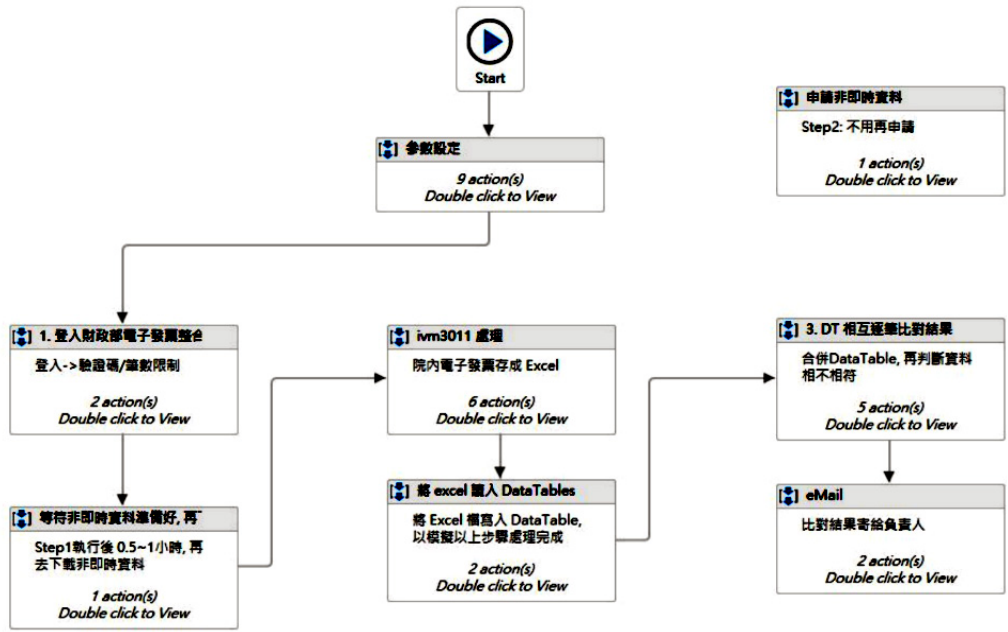

Fig. 2 The software robot for automating the audit of cloud E-invoice data. (in traditional Chinese)

Taiwan's E-invoice platform and the software robot shown in Figure 2 can eliminate the need for paper invoices or hard copies of E-invoices. Manufacturers only exchange E-invoice data with the Chi Mei Medical Center. This exchange may be through an email or a massager.

## 4. Results

The Chi Mei Medical Center had no confidence in a software robot since it was a young information technology. Therefore, they measured the average time in completing its conventional audit of paper invoices or hard copies of E-invoices at its three branches between October 2020 and January 2021. This average time is equal to

$$\text{average time} = \frac{\text{Total audit time}}{\text{Number of invoices}} \quad (1)$$

Comparing the results of Equation (1) with the average time spent to audit cloud E-invoice data using the proposed bot is next inspected.

Figs. 3(a)-3(b) denote the remote cause for introducing electronic invoicing. In these two figures, the Chi Mei Medical Center checks the total number of paper invoices and hard copies of E-invoices between October 2020 and January 2022. Its internal auditors may audit invoices one by one or in batches.

If the Chi Mei Medical Center kept auditing paper invoices and hard copies of E-invoices manually, it should maintain a big team of internal auditors to complete this task. Observing Figs. 3(a)-3(b) finds that hard copies of E-invoices increased gradually. Summing data points in these two figures reveals that the Chi Mei Medical Center audited 114214 invoices in batches and 11374 invoices one by one. Even if auditing invoices in batches, completing the audit spends unacceptably long time.

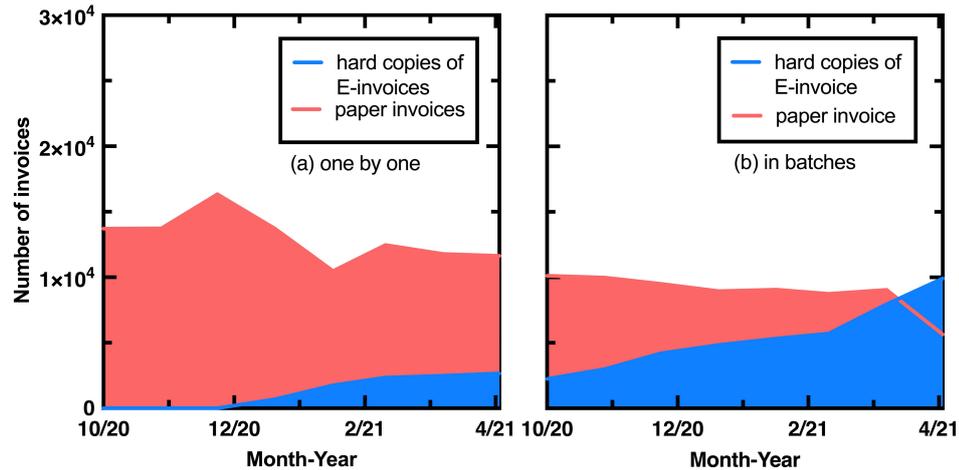

Fig. 3 Variation of invoices audited one-by-one or in batches: (a) one by one; (b) in batches

Furthermore, the executives of the Chi Mei Medical Center asked its internal auditors to sample some orders and measure the average time spent to audit invoices of these orders one by one. Placing the orders employs paper invoices or hard copies of E-invoices. Figs. 4(a)-4(c) show the results. These results serve as a benchmark for preliminarily evaluating the effects of electronic invoicing. Generating Figs. 4(a)-4(c) were completed at three branches of Chi Mei Medical Center from May 28, 2021, to Nov 4, 2021.

Even if the Chi Mei Medical Center just changed to replace some paper invoices with hard copies of E-invoices, Figs. 4(a)-4(c) demonstrate that auditing invoices one by one has been more efficient. In Figure 4(b), blue points generated with hard copies of E-invoices are below red points. These red points represent the conventional use of paper invoices. To internal auditors, the burden of auditing paper invoices was heavy. The average time spent to audit per paper invoice is about or above 1 minute. If the internal auditors of the Chi Mei Medical Center were few, they may need quite a while to audit over 1000 paper invoices one by one.

Nevertheless, Figs. 4(a)-4(c) contain too few data points. For example, observing Figs. 4(a) and 4(c) can find two blue points in these figures. Therefore, we still further question the efficiency of introducing electronic invoicing.

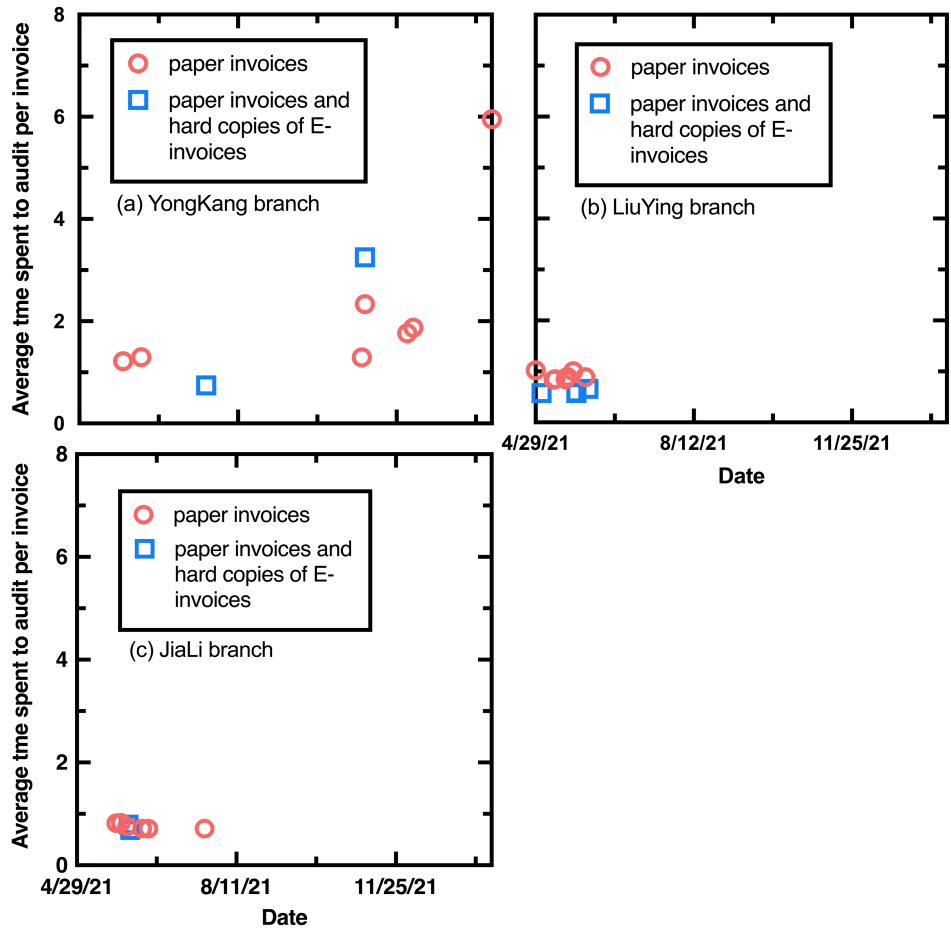

Figs. 4 Measurements of the average time spent to audit invoices one-by-one

Executives of the Chi Mei Medical Center also thought that Figs. 4(a)-4(c) were insufficient to conclude the benefits of electronic invoicing. They desired more evidence. Therefore, the Chi Mei Medical Center experimented to audit invoices in batch at its YongKang branch with paper invoices, hard copies of E-invoices, and cloud E-invoice data. The software robot shown in Fig. 2 was employed to audit cloud E-invoice data. The experiment time was between 10/18/2021 and 2/8/2021. For approaching the real working condition as close as possible, internal auditors of this YongKang branch may audit paper invoices, hard copies of E-invoices, and cloud E-invoice data simultaneously. Note that some manufacturers are old, they still use paper invoices. Figs. 5(a)-5(b) show the results.

Blue data points in Fig. 5(a) represent that replacing some paper invoices or hard copies of E-invoices with cloud E-invoice data improves the average time spent to audit invoices. Fig. 5(b) shows the total number of audited invoices increased after introducing electronic invoicing.

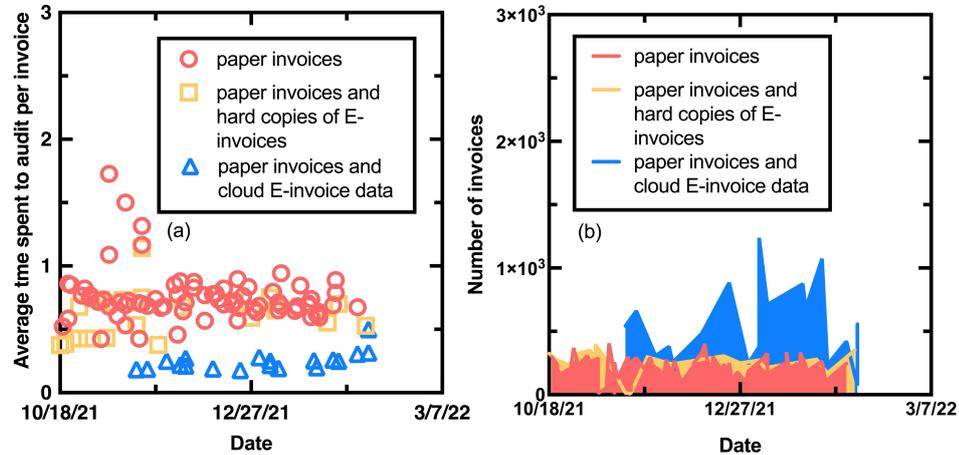

Fig. 5 Experiment of introducing electronic invoicing (audit of cloud E-invoice data in batch)

Figs. 6(a)-6(c) and Table 1 ensure that cloud E-invoice data and the proposed software robot reach the Chi Mei Medical Center's desire. In these three figures, three branches of Chi Mei Medical demonstrated that saving auditing time is possible if cloud E-invoice data and the proposed bot are adopted. Table 1 records that the Chi Mei Medical Center eliminated 9148 paper invoices in invoicing. Saving this amount brings profits to the Chi Mei Medical Center.

One may suspect that Table 1 didn't include computer usage, software fees, telecom charges, and electricity bills in operating the present software robot. Employing the UiPath (community version) can be free. Running this software requires a usual computer. In comparison with other uses such as healthcare in a hospital, a computer for operating the UiPath and a software robot is insignificant. In Taiwan, internet fees are fixed per month. Even if the Chi Mei Medical Center has not introduced a bot, it still pays fixed internet fees each month. Meanwhile, the electricity bills for employing the proposed software robot are negligible compared to the ordinary electricity bills of the Chi Mei Medical Center.

Internal auditors satisfied with the new electronic invoicing. One of the female internal auditors stated representative opinions of the proposed software robot:

"Oh! The software robot brings unexpected changes to my work. Now, I drink my bubble tea and wait for the outputs produced by the bot. My remaining work is confirming the results of this Excel file. That's great!"

The above conversation implies that existing internal auditors welcome a software robot. The vice president of the Chi Mei Medical Center was also surprised by the influence brought by software robots. He can't wait to see the application of other bots.

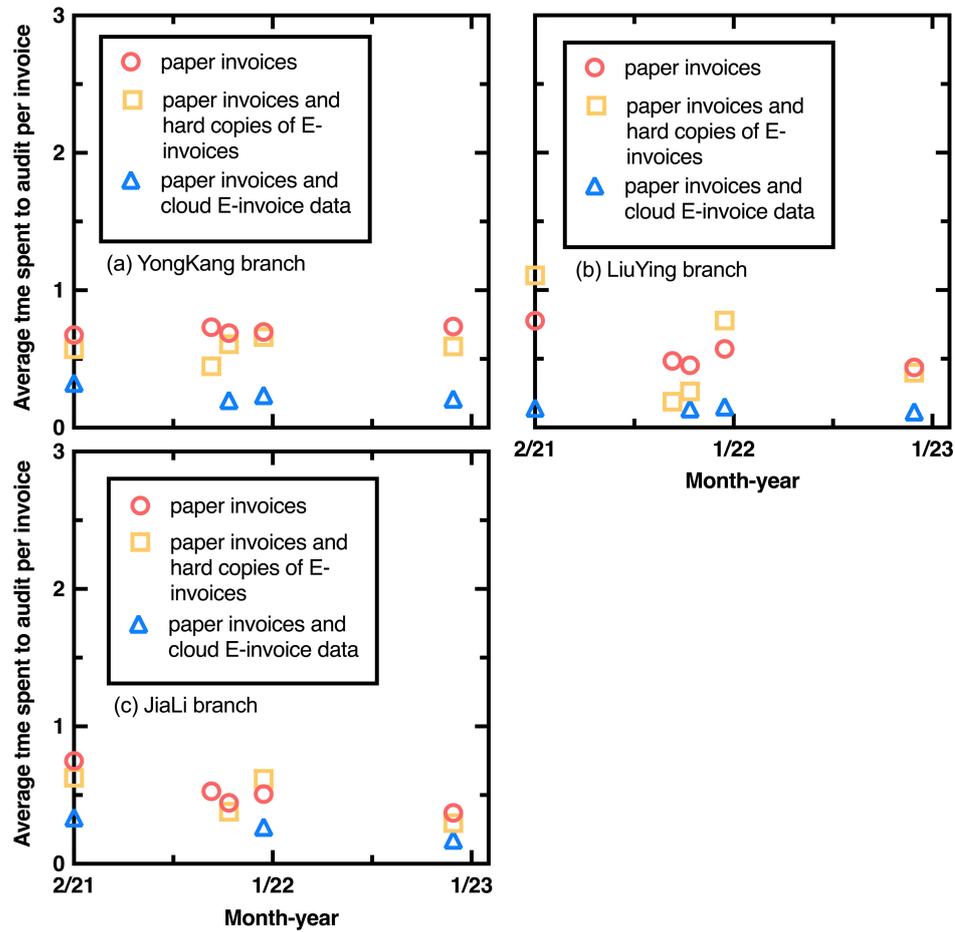

Fig. 6 Effects of promoting electronic invoicing to three branches of the Chi Mei Medical Center

Table 1 Changes before and after introducing the cloud electronic invoicing

| Observation | After | Before |
|---|---|---|
| paper invoices | 2,730 | 11878 |
| costs of papers | TWD 2,048 | TWD 10,409 |
| human resource costs | TWD 1,311,876 | TWD 1,457,640 |

## 4. Discussion

The Chi Mei Medical Center has completed four challenges concluded in published RPA researches [5]. The first challenge attributes to errors caused by scanned documents. That previous research concluded that an amount worth 1275.25 may be erroneously read as 1276.26 by a software robot. The Chi Mei Medical Center broke this challenge by integrating a software robot with cloud computing and QR codes. E-invoice data are stored in a website. The software robot operated including importing E-invoice data from this website.

The second challenge is the costs of deploying a software robot. The Chi Mei Medical Center saved money after deploying a software robot. This situation is different from other previous RPA researches which stated that we should be anxious about costs of building and deploying a software robot.

The third challenge is the resistance of human workers. The Chi Mei Medical Center operated their software robot when their internal auditors were off their office; therefore, the resistance of these internal auditors to the deployment of a software robot is mitigated. They demonstrated a good human-robot collaboration.

The four challenge is automating a dynamic process. As mentioned earlier, E-invoice may be uploaded immediately or periodically. The Chi Mei Medical Center properly arrange the logic of their software robot to wait new E-invoice data.

## 5. Conclusion

Some organizations hesitated to implement the RPA to improve their business processes since published RPA researches advise challenges such as high operation costs, resistance of human workers, and data errors provided by software robots. The Chi Mei Medical Center broke study these challenges. It combined the cloud computing, QR codes, and a software robot to implement the electronic invoicing; therefore, the possible errors produced by software robots were mitigated. Building the software robot spent as few costs as possible; hence, automating the audit of E-invoices using a bot was a reasonable investment. Internal auditors of the Chi Mei Medical Center showed an excellent human-robot collaboration. They welcomed the software bot since it simplified a tedious task. From the Chi Mei Medical Center's empirical evidence obtained in implementing electronic invoicing, this study concluded:
1. Successful deployment of a complex software robot requires the integration of other emerging technologies.
2. Mitigating the resistance of human workers to the RPA technology depends upon their human needs.

3. Current costs of creating and deploying software robots have been lower than the benefits obtained from using these bots.